\documentclass[conference]{IEEEtran}
\IEEEoverridecommandlockouts
\usepackage{cite}
\usepackage{multirow}
\usepackage{subcaption}
\usepackage{amsmath,amssymb,amsfonts}
\usepackage{algorithmic}
\usepackage{graphicx}
\usepackage{textcomp}
\usepackage{tabularx}
\usepackage{xcolor}
\def\BibTeX{{\rm B\kern-.05em{\sc i\kern-.025em b}\kern-.08em
    T\kern-.1667em\lower.7ex\hbox{E}\kern-.125emX}}
\begin{document}

\title{Comparison between the Structures of Word Co-occurrence and Word Similarity Networks for Ill-formed and Well-formed Texts in Taiwan Mandarin}

\author{\IEEEauthorblockN{1\textsuperscript{st} Po-Hsuan Huang}
\IEEEauthorblockA{
\textit{Department of Linguistics} \\
\textit{University of Southern California}\\
CA, U.S. \\
pohsuan@usc.edu
}
\and
\IEEEauthorblockN{2\textsuperscript{nd} Hsuan-Lei Shao}
\IEEEauthorblockA{\textit{Graduate Institute of Health and Biotechnology Law} \\
\textit{Taipei Medical University}\\
Taipei, Taiwan \\
Corresponding Author \\
ORCID:0000-0002-7101-5272}
}

\maketitle

\begin{abstract}
The study of word co-occurrence networks has attracted the attention of researchers due to their potential significance as well as applications. Understanding the structure of word co-occurrence networks is therefore important to fully realize their significance and usages. In past studies, word co-occurrence networks built on well-formed texts have been found to possess certain characteristics, including being small-world, following a two-regime power law distribution, and being generally disassortative. On the flip side, past studies have found that word co-occurrence networks built from ill-formed texts such as microblog posts may behave differently from those built from well-formed documents. While both kinds of word co-occurrence networks are small-world and disassortative, word co-occurrence networks built from ill-formed texts are scale-free and follow the power law distribution instead of the two-regime power law distribution. However, since past studies on the behaviour of word co-occurrence networks built from ill-formed texts only investigated English, the universality of such characteristics remains to be seen among different languages. In addition, it is yet to be investigated whether there could be possible similitude/differences between word co-occurrence networks and other potentially comparable networks. This study therefore investigates and compares the structure of word co-occurrence networks and word similarity networks based on Taiwan Mandarin ill-formed internet forum posts and compare them with those built with well-formed judicial judgments, and seeks to find out whether the three aforementioned properties (scale-free, small-world, and disassortative) for ill-formed and well-formed texts are universal among different languages and between word co-occurrence and word similarity networks.
\end{abstract}

%small_world:\cite{MasucciRodgers2006, Kramer2021}
%\cite{KapustinJamsen2007}
%\cite{MillingtonLuz2021}


\section{Introduction}
Word co-occurrence networks (WCN) have attracted the attention of researchers due to both their potential significance (e.g., semantic similarity\cite{Lancia2007}) as well as their applications (e.g., keyword extraction, text summarization, and author affiliation, cf. \cite{GargKumar2018}). The understanding of the structure of WCN is therefore crucial if one is to grasp a holistic picture of their significance and applications. Moreover, it is equally important to explore the possible similitudes/differences between WCN and other potentially comparable networks. In this study, we investigate and compare WCN and word similarity networks (WSN) for a Taiwan Mandarin internet forum, PTT and for judicial judgments made by Taiwanese courts from the years 2004 and 2008. In past studies, WCN based on well-formed documents have been found to share certain properties, including being small-world\cite{MasucciRodgers2006, Kramer2021}, following a two-regime power law distribution\cite{KapustinJamsen2007}, and being generally disassortative\cite{MillingtonLuz2021}. On the other hand, WCN built with less well-formed microblog data in English has been found to behave differently then WCN based on well-formed documents. While both kinds of WCN are small-world and disassortative, word co-occurrence networks built from ill-formed texts are scale-free and follow the power law distribution instead of the two-regime power law distribution\cite{GargKumar2018}. However, whether such likeness and discrepancies between WCN for well-formed and ill-formed texts can be universally found across languages requires further investigation. In addition, it remains to be seen whether such similarities and differences are reserved for WCN or are in fact shared among other networks such as networks based on word similarity. As such, the current study seeks to examine 1) the universality of the similarities and differences for the three parameters (degree distribution, small-wordness, and disassortativity) among different languages and 2) the universality of the three properties between WCN and WSN.

\section{Methods}

\subsection{Data collection}

For the PTT data, 139,578 posts from the \textit{Gossiping}, \textit{Food}, and \textit{HatePolitics} forums on PTT were collected between Jan. 1st to Jul. 31st, 2023. With the comments included, the dataset contained a total of 4,148,879 texts. For the judicial judgment data, 53,272 judgments during the years 2004 and 2008 were collected from the Judicial Yuan, R.O.C. Sentences were segmented with punctuation, leading to a total of 4,017,811 texts.

\subsection{Data preprocessing}
The texts were first preprocessed, with the numbers converted to \textit{0}, the alphabets converted into the lower cases, and all other non-Mandarin characters removed. The preprocessed texts were then segmented with the CKIP segmentation system\cite{TsaiChen2004}.

\subsection{Word embedding}
Word similarities were obtained with word embedding. A skip-gram word2vec model was first trained for the PTT data and judicial judgment data respectively, with a window size of 10. The vector sizes were 500. Only words with a occurrence more than 3 were taken into the vocabularies.

\subsection{Network building}
Four networks were built. For the PTT and judicial judgment data respecively, one unweighted and undirected networks were built based on word co-occurrence and word similarity respectively. These four networks were thus: a WCN and WSN for the PTT data (WCN-P and WSN-P) and a WCN and WSN for the judicial judgment data (WCN-J and WSN-J). To reduce the computational load, only 1/10 of the vocabulary was randomly selected as the database used for network building. For WCN, in the current study, the word co-occurrence was determined as two words that both occurred in the same text. For WSN, the similarity threshold was determined as the 99\textsuperscript{th} centile, and two words were determined as similar if the similarity was above the threshold. The numbers of the nodes and edges in these four networks are listed in Table \ref{table: Numbers of nodes and edges in the four networks.}.

\begin{table}[htbp]
\caption{Numbers of the nodes and edges in the four networks.}
\begin{center}
\begin{tabularx}{\columnwidth}{|X|X|X|}
\hline
\textbf{Network}&\textbf{Number of nodes}&\textbf{Number of edges} \\
\hline
WCN-P&24,035&208,759\\
\hline
WSN-P&228,163&43,106,445\\
\hline
WCN-J&16,847&1,469,475\\
\hline
WSN-J&86,262&43,248,161\\
\hline
\end{tabularx}
\label{table: Numbers of nodes and edges in the four networks.}
\end{center}
\end{table}

\subsection{Calculation of the parameters}

\subsubsection{Degree distribution}
The degree distributions of the four networks were accessed by fitting the degree distributions of the networks to power law vs. two-regime power law models. The models were then assessed with goodness of fit using the sums of squared residuals (SSR) and AIC.

\subsubsection{Small-world property}
The small-world property was determined by comparing the average clustering coefficients (CC) of the target network and the Erdos-Renyi (ER) random network\cite{ErdösRény2006}. A network is said to possess the small-world property if CC in the target network is far larger than CC in the ER random network ($\approx$0; cf.\cite{GargKumar2018}).

\subsubsection{Assortativity}
Lastly, the assortativity was decided with the degree assortativity coefficient (DAC), which indicated the tendency for a node to be connected with nodes with higher/lower degrees. If DAC is positive, the network is said to be assortative; if DAC is negative, the network is said to be disassortative. If the coefficient is close to zero, the network is said to be more randomly distributed in terms of its assortativity\cite{GargKumar2018}.

\section{Results}
% The WC and WS networks are visualized in Fig. \ref{figure: Visualization of WCN.} and Fig. \ref{figure: Visualization of WSN.}

% \begin{figure}[htbp]
% \centerline{
% \includegraphics[width=\columnwidth]{coocurrence_map.png}
% }
% \caption{Visualization of WCN.}
% \label{figure: Visualization of WCN.}
% \end{figure}
% \begin{figure}[htbp]
% \centerline{
% \includegraphics[width=\columnwidth]{similarity_map.png}
% }
% \caption{Visualization of WSN.}
% \label{figure: Visualization of WSN.}
% \end{figure}

\subsection{Degree distribution}
\begin{figure}[hbt!]
\begin{subfigure}[b]{\columnwidth}
\centerline{
\includegraphics[width=\columnwidth]{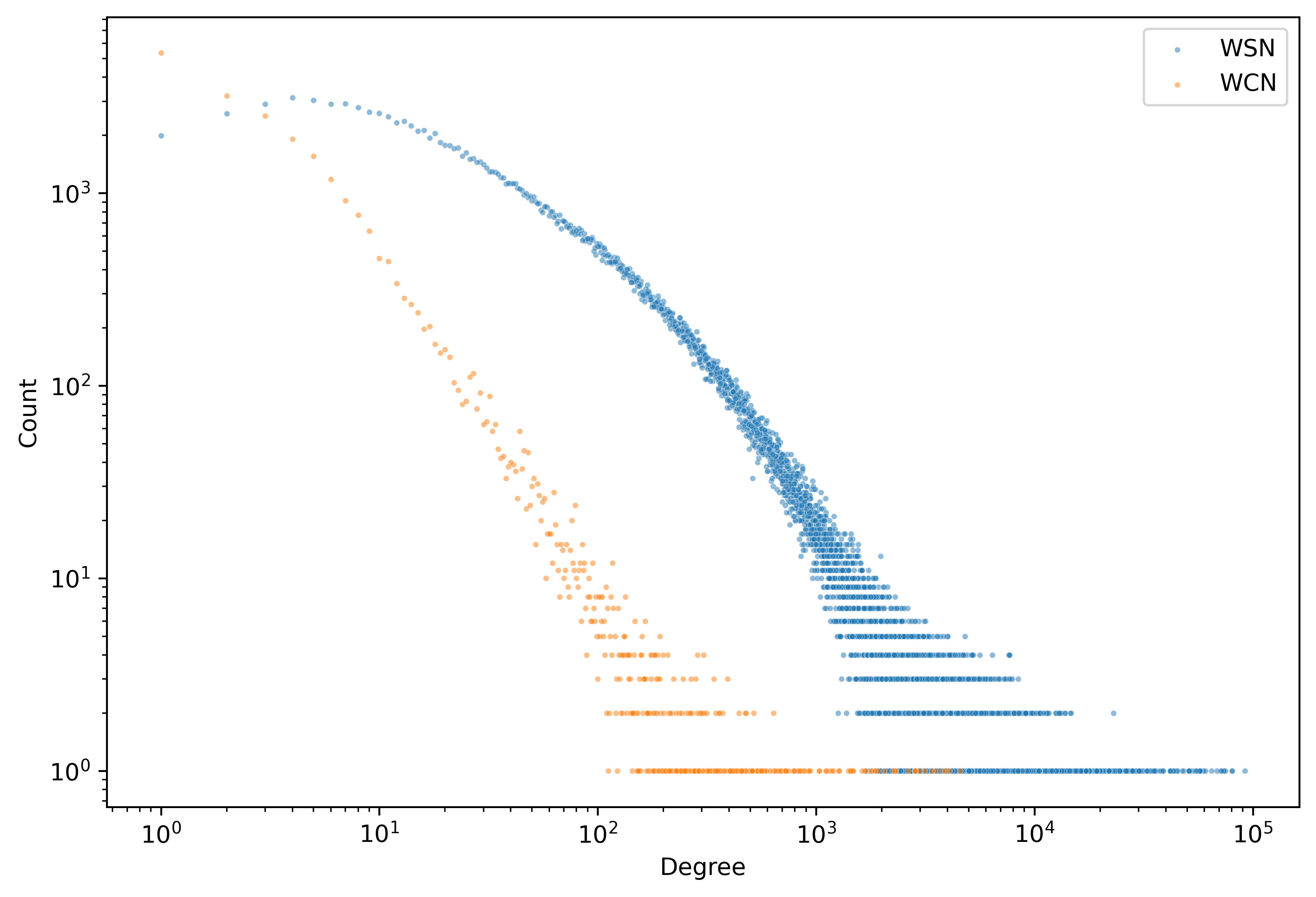}
}
\caption{PTT data}
\end{subfigure}
\begin{subfigure}[b]{\columnwidth}
\centerline{
\includegraphics[width=\columnwidth]{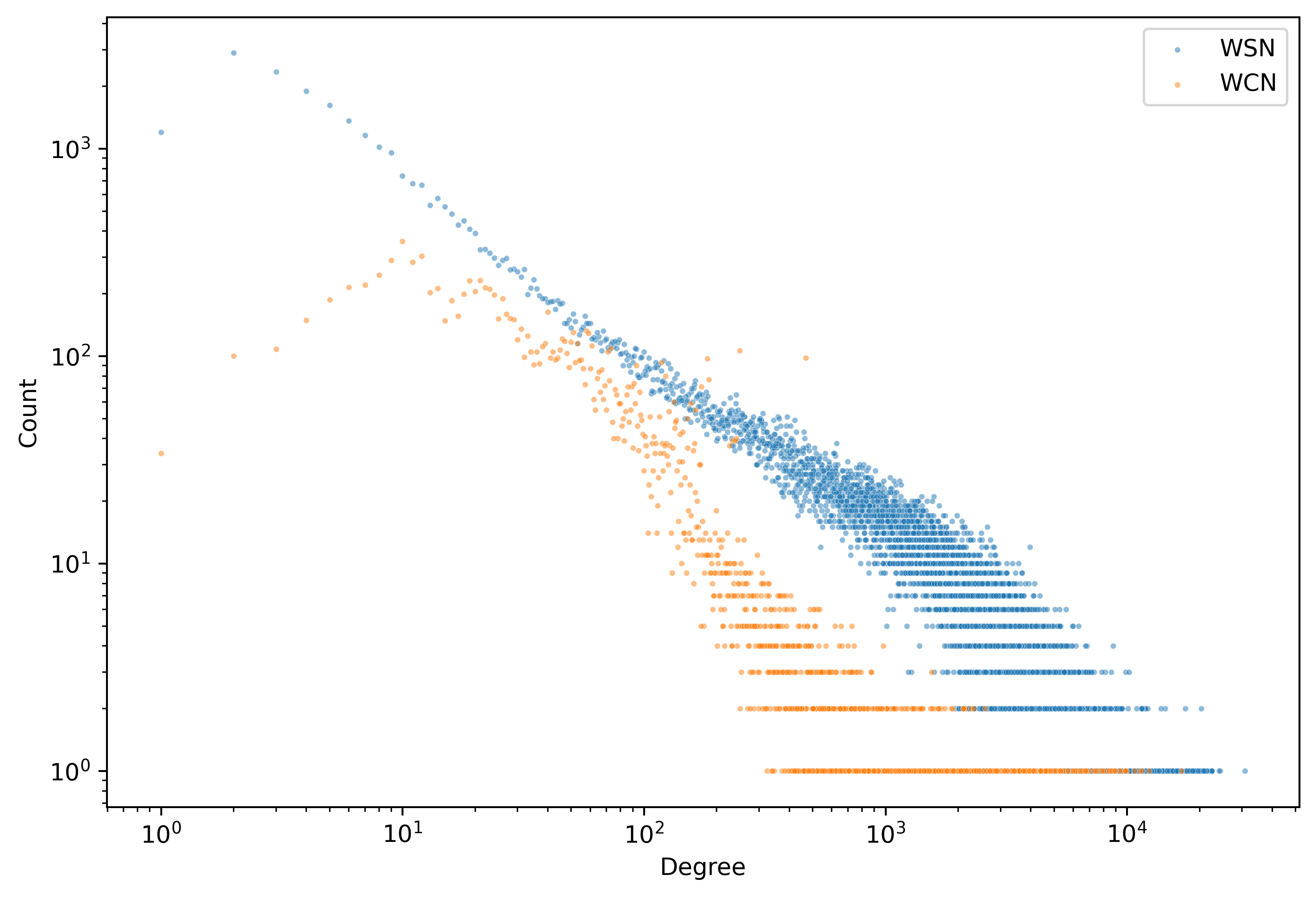}
}
\caption{Judicial judgment data}
\end{subfigure}
\caption{Degree distributions of word co-occurrence and similarity networks for the PTT and judicial judgment data.
}
\label{Figure: Degree distributions of word co-occurrence and similarity networks for the PTT and judicial judgment data.}
\end{figure}

The degree distributions of the four networks are illustrated in Figure \ref{Figure: Degree distributions of word co-occurrence and similarity networks for the PTT and judicial judgment data.}.

It can be observed that while WCN-P seems to be quite straight-forwardly scale-free, as is found in past studies\cite{GargKumar2018} for ill-formed texts, the other three networks seem to be rather ambiguous between being more similar to the power law distribution or to the two-regime distribution. However, upon exmination, our goodness-of-fit reults showed that all four networks were generally scale-free. As seen in Table \ref{table: Sums of squared residuals and AIC for fitted power-law and two-regime-power-law models for the four networks' degree distributions.}, for all four networks, the SSR were lower for the fitted power law models than the fitted two-regime power law models, and the AIC were also all lower for the fitted power law models than the fitted two-regime power law models.

\begin{table}[htbp]
\caption{Sums of squared residuals and AIC for fitted power-law and two-regime-power-law models for the four networks' degree distributions.}
\begin{center}
\begin{tabularx}{\columnwidth}{|l|X|l|l|}
\hline
\textbf{Network}&\textbf{Model}&\textbf{SSR}&\textbf{AIC}\\
\hline
\multirow{2}{*}{WCN-P}&power law&201.09&-284.67\\
\cline{2-4}
&two-regime power law&263.91&-169.74\\
\hline
\multirow{2}{*}{WSN-P}&power law&4,199.72&-2,760.59\\
\cline{2-4}
&two-regime power law&4,910.08&-1,748.96\\
\hline
\multirow{2}{*}{WCN-J}&power law&558.69&-1,000.83\\
\cline{2-4}
&two-regime power law&667.51&-774.55\\
\hline
\multirow{2}{*}{WSN-J}&power law&3,660.20&-4,321.66\\
\cline{2-4}
&two-regime power law&4,395.24&-3,060.43\\
\hline
\end{tabularx}
\label{table: Sums of squared residuals and AIC for fitted power-law and two-regime-power-law models for the four networks' degree distributions.}
\end{center}
\end{table}

\subsection{Small-worldness}
The clustering coefficients for the four networks are listed in Table \ref{table: Clustering coefficients for the four networks.}.

\begin{table}[htbp]
\caption{Clustering coefficients for the four networks.}
\begin{center}
\begin{tabularx}{.33\columnwidth}{|X|X|}
\hline
\textbf{Network}&\textbf{CC}\\
\hline
WCN-P&0.60\\
\hline
WSN-P&0.32\\
\hline
WCN-J&0.95\\
\hline
WSN-J&0.58\\
\hline
\end{tabularx}
\label{table: Clustering coefficients for the four networks.}
\end{center}
\end{table}

It can be seen that all four networks had clustering coefficients far larger than that of the ER random network, suggesting that all four models were small-world in nature.

\subsection{Assortativity}
The degree assortativity coefficients for the four networks are listed in Table \ref{table: Degree assortativity coefficients for the four networks.}.

\begin{table}[htbp]
\caption{Degree assortativity coefficients for the four networks.}
\begin{center}
\begin{tabularx}{.33\columnwidth}{|X|X|}
\hline
\textbf{Network}&\textbf{DAC}\\
\hline
WCN-P&-0.27\\
\hline
WSN-P&-0.18\\
\hline
WCN-J&-0.33\\
\hline
WSN-J&-0.04\\
\hline
\end{tabularx}
\label{table: Degree assortativity coefficients for the four networks.}
\end{center}
\end{table}

The negative DAC of the four networks suggest general disassortativity. However, the near zero value of the DAC for WSN-J suggests that WSN-J is rather neutral in terms of assortativity compared with the other three networks. In addition, it can also be observed that in general, WCN networks are more disassortative than WSN networks, for both well-formed and ill-formed data.

The three investigated parameters are summarized in Table \ref{table: Summary of the three parameters for the four networks.}.

\begin{table}[htbp]
\caption{Summary of the three parameters for the four networks.}
\begin{center}
\begin{tabularx}{\columnwidth}{|l|X|l|l|}
\hline
\textbf{Network}&\textbf{Degree distribution}&\textbf{Small-worldness}&\textbf{Assortativity}\\
\hline
WCN-P&\multirow{4}{*}{scale-free}&\multirow{4}{*}{small world}&\multirow{3}{*}{disassortative}\\
\cline{1-1}
WSN-P&&&\\
\cline{1-1}
WCN-J&&&\\
\cline{1-1}\cline{4-4}
WSN-J&&&Neutral\\
\hline
\end{tabularx}
\label{table: Summary of the three parameters for the four networks.}
\end{center}
\end{table}

\section{Discussion}
\subsection{Degree distribution for networks based on ill-formed and well-formed data}
As mentioned in \cite{GargKumar2018}, past studies of WCN built with well-formed texts generally fit two-regime power law distributions. On the other hand, their WCN built with Twitter microblog data is scale-free, and follows the power law distribution. However, our results suggest that such scale-free property may not be reserved to networks for ill-formed texts. Our networks for judicial judgment data also showed scale-free properties. Since judicial judgments are undoubtedly well-formed, such properties might not be directly related to the well-formedness of the texts, but rather may be attributed to the specificity of the texts. Since similar power law distribution has been found for other specific texts such as bioinformatics literature \cite{Lietal2018}, it is likely that in both ill-formed microblog texts such as Twitter and PTT data and specific texts such judicial judgment and academic literature data, a handful of specific words are connected to especially large numbers of words. In microblogs, such words may be acronyms or community-specific pronouns reserved for the community only. In specific texts, such words may be professional jargon. The exact determinant for such scale-free property, however, would require further investigation.

\subsection{Universality of the three parameters}
The results of the analysis for the three parameters for both word co-occurrence networks and word similarity networks based on Taiwan Mandarin texts showed that the two networks had similar structures as the word co-occurrence networks built with ill-fomred data as well as academic literature data in English in previous studies\cite{GargKumar2018, Lietal2018}. This indicates that these characteristics of networks for ill-formed/specific texts are universal across languages and are potentially shared among different networks.

\subsection{Potential different tendencies of word similarity networks}

While both the word co-occurrence networks and the word similarity networks investigated in the current study demonstrated similar properties. Different tendencies seemed to exist between the two kinds of networks. It can be noticed that, for the measured parameters, WSN seemed to behave differently from WCN invetigated in previous studies. The WSN in the current study had smaller CC and also DAC closer to 0, suggesting that their small-wordness was less significant than that of WCN, and that they were also less disassortative than WCN. However, it remains to be seen whether such discrepqncies between WCN and WSN can be extended to more balanced corpus or are present only when based on microblogs or specific texts.

\bibliographystyle{IEEEtran}
\bibliography{references.bib}

\end{document}